\documentclass[conference]{IEEEtran}
\IEEEoverridecommandlockouts

\usepackage{cite}
\usepackage{amsmath,amssymb,amsfonts}
\usepackage{algorithmic}
\usepackage{graphicx}
\usepackage{textcomp}
\usepackage{booktabs}
\usepackage{multirow}
\usepackage{xcolor}
\def\BibTeX{{\rm B\kern-.05em{\sc i\kern-.025em b}\kern-.08em
    T\kern-.1667em\lower.7ex\hbox{E}\kern-.125emX}}
\begin{document}

\title{Effective Resistance and Graph Neural Network Reliability in
Tissue-Specific Interactomes}

\author{\IEEEauthorblockN{Jianru Shen}
\IEEEauthorblockA{
\textit{University of Montana}\\
Missoula, MT, USA \\
js258133@umconnect.umt.edu}}

\maketitle

\begin{abstract}
Protein function annotation needs to know which predictions to distrust, not
only what a model predicts. We ask whether tissue-specific interaction
structure carries that information. Our candidate signal is effective
resistance, used previously to relieve over-squashing by rewiring. Across 24
tissue-specific interactomes it is dominated by inverse degree, and the
degeneration deepens as the co-expression filtered network grows, with a
Spearman correlation of $-0.955$. The residual departure from that limit
exceeds degree-preserving null graphs in all 24 networks. Controlling for
predictive entropy, degree, annotation cardinality, local structure and
feature-only difficulty, the residual explains additional per-node loss in 19
of 24 held-out networks once a permutation floor is subtracted, at every
depth, and the effect strengthens monotonically with depth. The increment
reaches
$0.37\%$ of the variance the controls leave
unexplained, $5.6$ times a permutation floor, against $1.5$ times when
the model is retrained in a degree-preserving null world. Selective
prediction improves negligibly. The signal is reproducible; degree
degeneration bounds it.
\end{abstract}


\begin{figure*}[t]
\centering
\includegraphics[width=\textwidth]{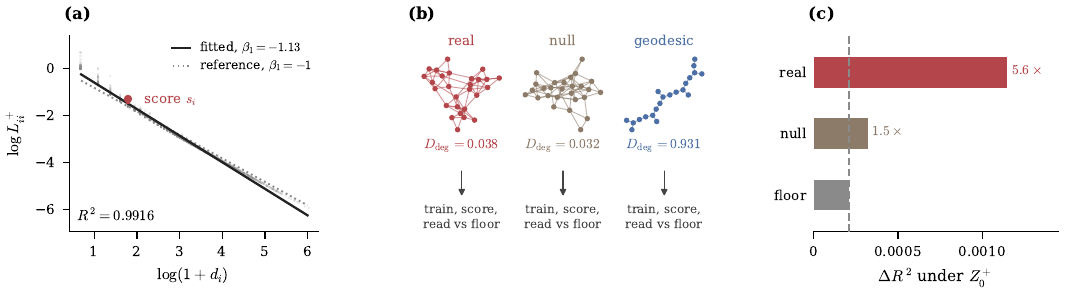}
\caption{Effective resistance in tissue-specific interactomes.
(a) On a representative network the log Laplacian pseudoinverse diagonal is
almost determined by log degree, and the structural score of a node is its
distance from the fitted line. The fitted slope is steeper than the reference
value of $-1$ that a pure inverse-degree relationship would give
asymptotically, which is why a fixed-coefficient correction leaves degree
dependence behind.
(b) The score is recomputed and the model retrained in three worlds: the real
interactome, a degree-preserving rewired null, and a near-tree geodesic
control where resistance reduces to graph distance. Networks are schematic
and the annotated values are corpus medians.
(c) Each increment is read against a permutation floor measured in its own
world, giving factors of $5.6$ and $1.5$ above that floor.}
\label{fig:overview}
\end{figure*}

\section{Introduction}

Protein function depends on tissue context, since the same protein
participates in different processes depending on where it is expressed
\cite{zitnik2017,greene2015}. Tissue-specific interactomes encode this by
restricting a global interaction network to proteins co-expressed in a given
tissue, and graph neural networks trained on them predict Gene Ontology terms
accurately \cite{hamilton2017,gligorijevic2021,radivojac2013}. Accuracy alone
does not
serve annotation pipelines. A curator needs to know which protein-level
predictions to distrust, because an incorrect functional label propagates
through every downstream analysis \cite{schnoes2009}.

Reliability estimates for these models come from the predictive distribution,
through entropy or Monte Carlo dropout \cite{gal2016,geifman2017,elyaniv2010}.
Such
scores use only the model output and ignore where in the interactome a
protein sits. Message passing aggregates along interaction paths, so a
structurally remote protein receives a degraded summary of the network, which
is what over-squashing describes \cite{alon2021,topping2022}. If structure
governs where information fails to arrive, it may also indicate where
predictions fail.

Effective resistance measures that remoteness. It is the resistance between
two nodes when each edge is a unit resistor \cite{klein1993}, it equals
commute time up to a constant \cite{chandra1989}, and it bounds
over-squashing, which motivated its use for selecting rewiring edges
\cite{black2023,digiovanni2023}. Whether it identifies which nodes a trained
model gets wrong is untested. The obstacle is degeneration. In well-mixed
graphs the resistance between two nodes converges to the sum of their inverse
degrees, leaving no global information \cite{vonluxburg2014}. Whether
tissue-specific interactomes fall in that regime, and whether anything
survives it, are empirical questions that available finite-graph bounds do
not settle for networks containing low-degree nodes.

Fig.~\ref{fig:overview} summarises the study. We compute exact Laplacian
pseudoinverses for all 24 networks of the standard tissue-specific benchmark,
measure how far each sits from the inverse-degree limit, and test whether the
residual predicts per-node failure beyond predictive confidence and
elementary structure. The answer bounds what any
resistance-aware reliability method can achieve on this class of data. Our
contributions are as follows.

\begin{itemize}
\item We place tissue-specific interactomes in the inverse-degree regime and
show that the degeneration deepens with network size, using near-tree
molecular graphs as a contrasting geodesic regime.
\item We show that real interactomes retain more corrected resistance
variation than degree-preserving null graphs, in all 24 tissues.
\item We show that the corrected residual explains additional per-node loss
in 19 of 24 held-out networks once a permutation floor is subtracted, and at
every depth, after controlling for predictive entropy, degree, annotation
cardinality, local structure and feature-only difficulty.
\item We quantify the practical ceiling. The increment is small, it does not
improve selective prediction, and it is strongly attenuated in a
degree-preserving null world.
\end{itemize}

\section{Background}

\subsection{Effective resistance}

Let $G=(V,E)$ be a connected undirected graph on $n=|V|$ nodes with adjacency
matrix $A$, degree $d_i=\sum_j A_{ij}$, degree matrix $D=\mathrm{diag}(d_i)$
and Laplacian $L=D-A$. Since $L$ is singular we use its Moore-Penrose
pseudoinverse $L^{+}$. The effective resistance between nodes $i$ and $j$ is
\begin{equation}
R_{\mathrm{eff}}(i,j) = L^{+}_{ii} + L^{+}_{jj} - 2L^{+}_{ij},
\label{eq:reff}
\end{equation}
the resistance measured between them when every edge is a unit resistor
\cite{klein1993}. It equals commute time divided by $2|E|$
\cite{chandra1989}, and its sum over all unordered pairs is the Kirchhoff
index $n\,\mathrm{tr}(L^{+})$, where $\mathrm{tr}$ denotes the trace
\cite{mohar1993,gutman1996}. The same quantity is used to sample edges for
spectral sparsification \cite{spielman2008}.

Because $R_{\mathrm{eff}}(i,i)=0$, summing \eqref{eq:reff} over all $j\in V$
gives
\begin{equation}
\textstyle\sum_{j\in V} R_{\mathrm{eff}}(i,j)
= n L^{+}_{ii} + \mathrm{tr}(L^{+}).
\label{eq:nodesum}
\end{equation}
The second term is constant within a graph, so $L^{+}_{ii}$ orders the nodes
of $G$ exactly as their total resistance to the rest of the network does.
This diagonal has been studied as a spreading centrality
\cite{vanmieghem2017}. We take it as our node-level measure of structural
remoteness.

\subsection{Two limiting regimes}

Effective resistance is informative only when it is not determined by simpler
quantities, and there are two regimes in which it is.

The first is the inverse-degree limit. In well-mixed graphs
$R_{\mathrm{eff}}(i,j)$ converges to $1/d_i+1/d_j$, so it carries nothing
beyond the two degrees \cite{vonluxburg2014}. The same work bounds the
finite-graph deviation in terms of the spectral gap of the random walk matrix
$D^{-1}A$, the maximum edge weight, and the minimum degree $d_{\min}$. That
bound scales as $1/d_{\min}^{2}$ and is therefore uninformative for networks
containing degree-one nodes, which includes every network we study. The
extent of the degeneration must be measured rather than certified.

The second is the geodesic limit. On a tree the unique path between two nodes
places their resistances in series, so $R_{\mathrm{eff}}$ coincides exactly
with the shortest-path distance $d_{\mathrm{sp}}$ and reduces to a
shortest-path statistic.

Write $i\nsim j$ for non-adjacent node pairs and $\mathrm{med}$ for the
median. Following \cite{vonluxburg2014} we measure the departure from the
first limit by
\begin{equation}
D_{\deg}(G) = \mathop{\mathrm{med}}_{i \nsim j,\; i \neq j}
\frac{\left|R_{\mathrm{eff}}(i,j) - 1/d_i - 1/d_j\right|}
{R_{\mathrm{eff}}(i,j)},
\label{eq:ddeg}
\end{equation}
and the departure from the second by
\begin{equation}
D_{\mathrm{geo}}(G) = \mathop{\mathrm{med}}_{i \neq j}
\frac{\left|R_{\mathrm{eff}}(i,j) - d_{\mathrm{sp}}(i,j)\right|}
{d_{\mathrm{sp}}(i,j)}.
\label{eq:dgeo}
\end{equation}
Each deviation is normalised by the quantity that is exact in its own regime,
so the two denominators differ. Small $D_{\deg}$ places a graph near the
inverse-degree limit and small $D_{\mathrm{geo}}$ near the geodesic limit.
We restrict \eqref{eq:ddeg} to non-adjacent pairs because shorting all other
nodes together gives $R_{\mathrm{eff}}(i,j)\geq 1/d_i+1/d_j$ for such pairs
by Rayleigh monotonicity, so the numerator is a non-negative excess rather
than a signed residual. Unlike \cite{vonluxburg2014}, which subsamples node
pairs on large graphs, we evaluate both quantities over all pairs.

\subsection{Effective resistance and message passing}

Message passing compresses a growing neighbourhood into a fixed-width node
state, and over-squashing describes the resulting loss
\cite{alon2021,topping2022}. Total effective resistance bounds this loss
\cite{black2023}, and the pairs most affected are those at high commute time
\cite{digiovanni2023}. Both results were used to select edges to add so that
resistance falls, that is, to modify the graph before training. We use
resistance differently, as a fixed descriptor of a node in an unmodified
biological network, and ask whether it identifies where a trained model
fails. Depth matters here because \cite{digiovanni2023} shows that adding
layers does not relieve over-squashing but lets vanishing gradients dominate
it, which predicts that deeper models fail more systematically at
high-resistance nodes.

\section{Scores, Risk Models and Controls}

\subsection{Structural score}

Effective resistance must be stripped of its degenerate component before it
can be tested for incremental information. Within each graph we regress the
log diagonal on log degree,
\begin{equation}
\log L^{+}_{ii} = \beta_0 + \beta_1 \log(1+d_i) + \varepsilon_i ,
\label{eq:score}
\end{equation}
fit by ordinary least squares, and define the structural score $s_i$ as the
standardised residual $\hat\varepsilon_i$, centred and scaled within the
graph. The score is label free and model free, so it is available for any
node of any graph without reference to a trained model.

Fitting $\beta_1$ rather than fixing it is deliberate. The inverse-degree
limit predicts a coefficient of one on $1/d_i$, but the fitted slope departs
from that value on every network we study, so a fixed-coefficient correction
leaves degree dependence behind. We report the theory-aligned alternative
\begin{equation}
c_{\mathrm{ER}}(i) = L^{+}_{ii} - \frac{n-2}{n\,d_i}
\label{eq:cer}
\end{equation}
as a comparison quantity, and also a variant $s^{\mathrm{full}}_i$ obtained
by adding mean neighbour degree, local clustering coefficient and two-hop
neighbourhood size to the right side of \eqref{eq:score}.

\subsection{Risk target and nested models}

For a trained model let $\ell_i$ denote the per-node loss, the binary
cross-entropy of the predicted probabilities against the true label vector,
averaged over the label set. Let $H_i$ denote the predictive entropy of the
same probabilities, averaged over the label set, and let $a_i$ denote the
annotation cardinality, the number of positive labels at node $i$.

The question is whether $s_i$ explains $\ell_i$ beyond what confidence and
elementary structure already explain. We answer it by nested least squares.
Let $Z_0$ collect an intercept together with $H_i$, $\log(1+d_i)$ and $a_i$,
and let $Z_0^{+}$ extend $Z_0$ with $\log(1+b_i)$, $k_i$ and
$\log(1+\beta_i)$, where $b_i$ is the mean degree of the neighbours of $i$,
$k_i$ its local clustering coefficient and $\beta_i$ the number of nodes
within two hops. All columns are standardised within the graph. For a design
matrix $Z$ write
\begin{equation}
R^{2}(Z) = 1 - \frac{\lVert \ell - Z\hat\beta \rVert^{2}}
{\lVert \ell - \bar\ell\mathbf{1} \rVert^{2}},
\label{eq:r2}
\end{equation}
and define the incremental contribution of the score as
\begin{equation}
\Delta R^{2}(G) = R^{2}\!\left([\,Z_0^{+},\, s\,]\right) - R^{2}(Z_0^{+}).
\label{eq:dr2}
\end{equation}
The looser control $Z_0$ is reported alongside $Z_0^{+}$ so that the
contribution of local structure is visible. Predictive entropy belongs in
$Z_0$ because for a calibrated model \cite{guo2017} the expected loss at a node
equals the
entropy of its predicted distribution, which makes it the strongest available
baseline rather than a convenient one.

For the selective-prediction view we label a node at risk if $\ell_i$ exceeds
the within-graph $80$th percentile and fit a logistic model. Here the design
drops annotation cardinality, since that quantity needs the labels and is
unavailable when deciding whether to trust a prediction; every remaining
covariate is observable at inference. We report area under the
precision-recall curve together with area under the risk-coverage curve, the
latter computed by accepting nodes in increasing order of predicted risk.

\subsection{Permutation floor}

Adding any column to a least squares design cannot decrease $R^{2}$, so
\eqref{eq:dr2} is positive even for an uninformative score. We therefore
estimate a floor. Let $\pi$ be a uniformly random permutation of the node
indices of $G$ and let $s_{\pi(i)}$ replace $s_i$, which destroys the
correspondence between structure and loss while preserving the marginal
distribution of the score. Averaging over one thousand draws gives
$\Delta R^{2}_{\pi}(G)$, and $\Delta R^{2}(G) - \Delta R^{2}_{\pi}(G)$ is the
corrected increment. A thousand draws are needed rather than a handful: the
standard deviation of a single draw is $3.1\times 10^{-4}$, larger than the
floor itself, so a small sample cannot support a ratio quoted to two
significant figures. The floor also grows as the sample shrinks, so it is
re-estimated inside every subset for which an increment is reported.

\subsection{Three worlds}

The score is evaluated in three settings that differ in what structure is
available.

The \emph{real} setting uses the tissue-specific interactomes unchanged.

The \emph{degree-preserving null} replaces each interactome by a rewired
graph obtained through double-edge swaps, which leave every degree unchanged
while destroying the arrangement of edges. Structure and model are both moved
into this world: the score is recomputed on the rewired graph and the model
is trained, validated and tested there. Any increment that survives is a
property of the degree sequence alone. The comparison is scaled to each
world's own permutation floor, because the two worlds support models of
different quality and only the presence of the effect, not its magnitude, is
comparable across them.

The \emph{geodesic control} is a corpus of near-tree molecular graphs on
which resistance provably reduces to shortest-path distance. It tests
specificity: a pipeline that manufactures signal would report an effect where
theory says none exists.

\subsection{Unit of analysis}

Nodes within an interactome are not independent, so every quantity is
aggregated to one value per graph, averaging over training seeds, and
confidence intervals are obtained by resampling the graphs.

\begin{table*}[t]
\centering
\caption{Resistance regimes. Medians across graphs for the three families,
ranges for size and spectrum. Density does not set the regime: the barbell is
five times denser than the dense random graph yet sits at the opposite
corner, and its small $\lambda_2$ identifies the bottleneck responsible.}
\label{tab:regimes}
\footnotesize
\begin{tabular}{lrrrrrr}
\toprule
Graph family or anchor & Graphs & Nodes & Mean degree & $\lambda_2$ &
$D_{\deg}$ & $D_{\mathrm{geo}}$ \\
\midrule
Tissue-specific interactomes & 24 & 591--3480 & 13.0--32.1 & 0.16--0.53
& 0.038 & 0.931 \\
Degree-preserving null & 72 & 591--3480 & 13.0--32.1 & 0.30--0.85
& 0.032 & 0.931 \\
Peptides-func & 15535 & 5--444 & 1.6--2.2 & 0.000--0.519
& 0.931 & 0.000 \\
\midrule
Path & 1 & 300 & 1.99 & 0.0001 & 0.989 & 0.000 \\
Random tree & 1 & 300 & 1.99 & 0.0006 & 0.940 & 0.000 \\
Ring lattice, $k=4$ & 1 & 300 & 4.00 & 0.0022 & 0.957 & 0.697 \\
Barbell & 1 & 305 & 146.60 & 0.0022 & 0.998 & 0.247 \\
Stochastic block model, 4 blocks & 1 & 300 & 13.15 & 2.032 & 0.104 & 0.932 \\
Erd\H{o}s-R\'enyi, sparse & 1 & 299 & 6.00 & 0.336 & 0.191 & 0.859 \\
Erd\H{o}s-R\'enyi, dense & 1 & 300 & 30.09 & 13.971 & 0.031 & 0.965 \\
\bottomrule
\end{tabular}
\end{table*}
\section{Data and Protocol}

\subsection{Tissue-specific interactomes}

We use the 24 tissue-specific protein interaction networks of the standard
inductive benchmark \cite{hamilton2017}, derived from the multi-layer tissue
networks of \cite{zitnik2017}. Each is the subgraph of a global human
interactome induced by the proteins co-expressed in one tissue, an
interaction being retained when both proteins are tissue specific or one is
tissue specific and the other ubiquitous. Node features are 50 gene-set
memberships from the Molecular Signatures Database \cite{subramanian2005} and
the 121 labels are Gene Ontology biological process terms
\cite{ashburner2000}. Networks average 2373 nodes and degrees span 1 to 720.
Three networks are disconnected, so structural quantities are computed on the
largest connected component, which retains at least $87.5\%$ of nodes and all
of them in 21 networks.

The specificity control is Peptides-func \cite{dwivedi2022,tonshoff2024}, 15535
peptide
molecular graphs with mean degree 2.03 and a mean bridge fraction of 0.83.
These are trees plus a few side-chain rings, mean cycle rank 3.74 against
mean diameter 57.0. On the $9\%$ of the corpus that are exact trees,
effective resistance equals shortest-path distance identically; across the
whole corpus the median relative discrepancy between the two is $0.0103$. The
correct answer is therefore known in advance to within one percent.

Null graphs come from double-edge swaps \cite{maslov2002,milo2002}, $10|E|$
attempts per graph and three
replicas per interactome, all preserving connectivity. Swapping halves mean
clustering, from 0.182 to 0.090, rather than eliminating it, and removes
larger-scale arrangement at the same time. Results against this null
therefore support claims beyond the degree sequence and do not separate local
clustering from larger-scale organisation.

\subsection{Model and protocol}

The predictive model is GraphSAGE \cite{hamilton2017} with mean aggregation,
width 128, and 2, 4 or 6 message-passing layers with residual connections and
layer normalisation. Training runs 300 epochs of Adam \cite{kingma2015} under a
cosine schedule
from $10^{-2}$ to $10^{-5}$ with no early stopping, so convergence follows
from the schedule rather than a stopping rule; validation micro-F1 moves by
less than $0.001$ over the final 30 epochs at every depth. Predictive
quantities average 20 stochastic passes with dropout active. A feature-only
multilayer perceptron of matching width and depth controls for node
difficulty not arising from the graph.

Four outer folds of 15 training, 3 validation and 6 test networks hold out
every interactome exactly once, with five seeds per fold. This departs from
the conventional split, so micro-F1 is not comparable to published values.
The grid comprises 80 training runs, and repeating the experiment inside the
null world at the primary depth across three replicas adds 60 more.

Models are implemented with PyTorch Geometric \cite{fey2019}. All structural
quantities are exact. A dense eigendecomposition of the
largest interactome takes 1.91 seconds and the full pass over 24 networks
takes 47 seconds, so no spectral approximation is needed at this scale.
Appendix~\ref{app:numerics} verifies the computation against two independent
routes and against the analytic answer on trees.

\begin{table}[t]
\centering
\caption{Base model quality. Validation micro-F1 over four folds and five
seeds, with the standard deviation across runs.}
\label{tab:models}
\footnotesize
\begin{tabular}{lr}
\toprule
Model & Micro-F1 \\
\midrule
Feature-only perceptron & $0.5546 \pm 0.0038$ \\
GraphSAGE, 2 layers & $0.8399 \pm 0.0054$ \\
GraphSAGE, 4 layers & $0.8799 \pm 0.0055$ \\
GraphSAGE, 6 layers & $0.9026 \pm 0.0057$ \\
\midrule
GraphSAGE, 6 layers, null world & $0.5972 \pm 0.0037$ \\
\bottomrule
\end{tabular}
\end{table}

\begin{table*}[t]
\centering
\caption{Incremental variance in per-node loss from each structural score,
per graph, averaged over seeds, with $95\%$ confidence intervals from
resampling the 24 interactomes. All rows outside the depth block use six
message-passing layers. The permutation floor, estimated from one thousand
draws per graph, is the scale against which every other row should be read.
Since $\Delta R^{2}$ cannot be negative in sample, the count of positive
interactomes is informative only once that floor is subtracted, which leaves
19 of 24 for the primary score at six layers.}
\label{tab:increments}
\footnotesize
\setlength{\tabcolsep}{5pt}
\begin{tabular}{lccr}
\toprule
Score and setting & $\Delta R^{2}$ under $Z_0$ & $\Delta R^{2}$ under
$Z_0^{+}$ & Interactomes \\
\midrule
Permutation floor & --- & $+0.00021$ [$+0.00017$, $+0.00025$] & --- \\
\midrule
\multicolumn{4}{l}{\emph{Primary score $s$, by depth}} \\
\quad 2 layers & $+0.00237$ [$+0.00167$, $+0.00315$] & $+0.00084$ [$+0.00052$, $+0.00119$] & 24/24 \\
\quad 4 layers & $+0.00306$ [$+0.00225$, $+0.00400$] & $+0.00094$ [$+0.00062$, $+0.00132$] & 24/24 \\
\quad \textbf{6 layers} & $\mathbf{+0.00330}$ [$+0.00232$, $+0.00444$] & $\mathbf{+0.00115}$ [$+0.00074$, $+0.00164$] & 24/24 \\
\quad \emph{\quad floor subtracted} & --- & $+0.00095$ [$+0.00052$, $+0.00143$] & \textbf{19/24} \\
\midrule
\multicolumn{4}{l}{\emph{Alternative scores}} \\
\quad Theory-aligned $c_{\mathrm{ER}}$ & $+0.00199$ [$+0.00075$, $+0.00404$] & $+0.00106$ [$+0.00021$, $+0.00263$] & 24/24 \\
\quad Locally corrected $s^{\mathrm{full}}$ & $+0.00115$ [$+0.00075$, $+0.00163$] & $+0.00115$ [$+0.00075$, $+0.00165$] & 24/24 \\
\midrule
\multicolumn{4}{l}{\emph{Controls}} \\
\quad Perceptron loss in design & $+0.00339$ [$+0.00237$, $+0.00449$] & --- & 24/24 \\
\quad Degree-preserving null world & $+0.00201$ [$+0.00125$, $+0.00291$] & $+0.00032$ [$+0.00021$, $+0.00047$] & 24/24 \\
\bottomrule
\end{tabular}
\end{table*}

\section{Results}

\subsection{Interactomes lie in the inverse-degree regime}

Table~\ref{tab:regimes} places the two biological corpora at opposite corners
of the plane, a factor of $24$ apart on $D_{\deg}$. For interactomes the
typical departure of a resistance from $1/d_i+1/d_j$ is under one tenth of
its value. Peptide graphs have a median $D_{\mathrm{geo}}$ of exactly zero,
so for more than half of these molecules the resistance between two atoms
equals their graph distance. The anchors show that the plane is genuinely
two-dimensional rather than one curve: the path, the barbell and the ring
lattice all sit near $D_{\deg}=1$ while their $D_{\mathrm{geo}}$ ranges from
zero to $0.70$.

Degeneration deepens as the interactome grows. Across the 24 networks the
Spearman correlation between node count and $D_{\deg}$ is $-0.955$, and
$D_{\deg}$ falls from $0.085$ in the smallest network to $0.033$ in the
largest. Since these networks are subgraphs of a common interactome induced
by tissue co-expression, resistance carries progressively less structural
information as the co-expression filtered interaction context widens. The
finite-graph bound of \cite{vonluxburg2014} does not certify this, because it
scales as $1/d_{\min}^{2}$ and every network here contains degree-one nodes.

\subsection{Real interactomes exceed their degree-matched nulls}

\begin{figure}[t]
\centering
\includegraphics[width=0.85\columnwidth]{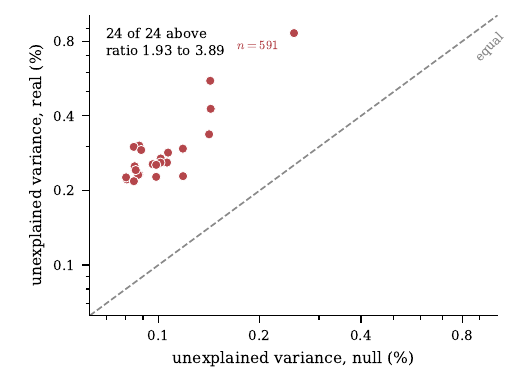}
\caption{Variance of the log pseudoinverse diagonal left unexplained after
degree, neighbour degree, clustering and two-hop neighbourhood size, for each
interactome against its own degree-preserving null. Every network lies above
the diagonal. The largest departure belongs to the smallest network, matching
the size trend in $D_{\deg}$.}
\label{fig:evidence}
\end{figure}

Fitting \eqref{eq:score} on real interactomes leaves $0.30\%$ of the variance
of $\log L^{+}_{ii}$ unexplained after neighbour degree, clustering and
two-hop neighbourhood size are also included, against $0.10\%$ on
degree-preserving nulls. The real network is higher in all 24 pairs, with
ratios from $1.93$ to $3.89$ and a mean of $2.80$, as
Fig.~\ref{fig:evidence} shows. 

The null perturbs the coordinate it is designed to perturb. Rewiring moves
$D_{\deg}$ by $+0.0078$ on average but $D_{\mathrm{geo}}$ by only
$+0.00005$, a factor of 150.

\subsection{The geodesic control behaves as theory requires}

On the 1412 peptide graphs that are exact trees the largest observed
discrepancy between resistance and graph distance is
$1.7 \times 10^{-10}$, which is numerical rather than structural. Over the
full corpus the median relative discrepancy is $0.0103$. Regressing mean
resistance on elementary geodesic summaries leaves $0.77\%$ of variance, of
which the cycle rank contributes $0.0002$ against $0.175$ for the diameter.
On near-tree molecules resistance is therefore governed by size and
elongation and not by ring count, and a pipeline that reported a strong
resistance effect here would be manufacturing it.

\subsection{The residual explains per-node loss}

Table~\ref{tab:models} establishes that the predictive models are
non-trivial. The six-layer network reaches $0.9026$ micro-F1 against $0.5546$
for a feature-only perceptron of matching capacity, so more than half of the
attainable signal comes from the interaction structure itself.
Validation micro-F1 changes by at most $0.0006$ over the final 30 epochs at
every depth, so the schedule rather than a stopping rule determines
convergence.

Table~\ref{tab:increments} and Fig.~\ref{fig:overview}c give the central
result. Under the strict design $Z_0^{+}$ the structural score adds
$+0.00115$ of explained variance in per-node loss, with a confidence interval
from $+0.00074$ to $+0.00164$. Since $\Delta R^{2}$ cannot be negative in
sample, the meaningful count is the 19 of 24 interactomes that stay positive
once each floor is subtracted. Two independent routes agree on the
$+0.00115$ estimate. Removing local structure from the score gives
$s^{\mathrm{full}}$ under $Z_0$, and adding it to the design gives $s$ under
$Z_0^{+}$; both yield $+0.00115$. Roughly two thirds of the looser increment
under $Z_0$ is therefore local structure and one third is not.

The direction is consistent with the over-squashing account. Across 120 fits
of one interactome and one seed each, the coefficient on the score is
$+0.00616$ with a confidence interval from $+0.00537$ to $+0.00697$ and is
positive in 113 of them, so structurally remote proteins carry higher loss.

\subsection{The effect strengthens with depth}

The increment rises monotonically from $+0.00084$ at two layers to
$+0.00115$ at six, positive in all 24 interactomes at every depth. The ratio
between six and two layers is $1.37$ under $Z_0^{+}$ and $1.39$ under $Z_0$,
so the depth sensitivity survives the strict control and resides in the
global component of the score rather than the local one. This matches the
prediction of \cite{digiovanni2023} that depth does not relieve
over-squashing but allows vanishing gradients to dominate it, which should
make deeper models fail more systematically at remote nodes.

\subsection{Controls}

Adding the per-node loss of the feature-only perceptron to $Z_0$ raises its
explained variance but leaves the increment unchanged at $+0.00339$, positive
in all 24 interactomes. The residual therefore explains GNN loss
beyond feature-only difficulty, although this alone does not identify a
message-passing-specific mechanism.

In the degree-preserving null world, where the model is retrained and the
score recomputed on rewired graphs, the increment falls to $+0.00032$. Read
against each world's own permutation floor, the real interactomes give a
factor of $5.60$ and the null world a factor of $1.50$, so in the null world
the score contributes little more than an uninformative variable. The paired
difference of floor-corrected increments is $+0.00084$ with a confidence
interval from $+0.00042$ to $+0.00130$ and favours the real network in 18 of
24 pairs. Per-graph consistency is 24 of 24 for the raw increment and 19 of
24 once each interactome's own floor is subtracted. The null world also costs
accuracy: the retrained model reaches $0.5972$ micro-F1, retaining only
$12.2\%$ of the advantage the graph confers over the feature-only control.

\subsection{A mechanism test that fails}

A node-level test of the mechanism fails. Two-hop coverage shows the expected
monotone trend, but it is collinear with degree at $\rho=0.917$ and degree is
already in $Z_0^{+}$. Blocking on degree removes the trend, giving a pooled
difference of $-0.00014$ $[-0.00110, +0.00102]$, positive in 31 of 72 cells,
and a continuous score-by-coverage interaction has the opposite sign to the
prediction. The mechanism evidence here is therefore model-level, through
depth, and not node-level, which is consistent with a gradient-scale effect
but does not demonstrate one. Appendix~\ref{app:receptive} gives the full
analysis.

\subsection{The practical ceiling}

The design $Z_0$ already explains $0.6113$ of the variance in per-node loss,
and the structural score claims $0.37\%$ of what the strict design leaves
unexplained. Selective prediction is evaluated on a deployable design that
drops annotation cardinality, so that every covariate is available before the
labels are known. Under it the area under the risk-coverage curve improves
from $0.0405$ to $0.0404$, in 22 of 24 interactomes, and the area under the
precision-recall curve from $0.7954$ to $0.7958$, a gain of $+0.00034$
$[+0.00016, +0.00053]$ that claims $0.17\%$ of the headroom those covariates
leave. The association is reproducible across every interactome, every depth
and both correction routes, and it is too small to change a deferral
decision. Effect size does not increase with network size either: the
Spearman correlation between node count and the increment is $0.349$ over 24
networks, a weak trend that the sample does not resolve.

\section{Discussion and Conclusion}

Effective resistance carries information about which protein-level
predictions a graph neural network will get wrong, and degree degeneration
bounds how much. Both halves are supported. The association survives a design
containing predictive entropy, degree, annotation cardinality, local
structure and feature-only loss; it holds at every depth and through two
independent correction routes; and its sign matches the over-squashing
account in 113 of 120 fits. It is also small enough to change no deferral
decision.

The ceiling has an identifiable cause. Resistance between two proteins here
is within a tenth of the sum of their inverse degrees, and that departure
shrinks as the co-expression filtered network grows. Most of what resistance
could contribute has already been absorbed by degree, which any reliability
model has for free. The room available to a resistance-aware node score on
tissue-specific interactomes is bounded by how far those networks sit from
the inverse-degree limit, and that distance is small.

\subsection{Scope}

We evaluate resistance as a standalone node-level risk score on one
benchmark, one architecture family and one uncertainty target. We do not
evaluate resistance-based rewiring, the setting in which resistance has been
shown to help \cite{black2023,digiovanni2023}; nothing here bears on it. A
quantity that is weak on its own may behave differently in networks further
from the degree limit, and Table~\ref{tab:regimes} shows that such networks
exist.

The work identifies no biological mechanism. Node identifiers are not
recoverable from the public release of this benchmark, so no protein, tissue
or pathway can be named. What can be said biologically is structural:
interaction contexts that retain more proteins after co-expression filtering
are more strongly degree-dominated, so global resistance describes them less.

\subsection{Limitations}

Rewiring halves clustering rather than eliminating it, so the null supports
claims beyond the degree sequence but does not separate local clustering from
larger-scale organisation. Per-graph consistency falls from 24 of 24 to 19 of
24 once each floor is subtracted, so the aggregate effect is more robust than
its per-network sign. The mechanism evidence is model-level only. The risk
experiment uses a single benchmark, since Peptides-func serves as a
specificity control rather than a second arm.

\subsection{Outlook}

Two directions follow. The first is whether the ceiling is a property of
resistance or of these networks, tested by repeating the measurement on
biological graphs further from the inverse-degree limit, where
Table~\ref{tab:regimes} predicts more room. The second is whether resistance
is more useful as a design quantity than a diagnostic one, tested by
comparing the ceiling measured here against the gain from rewiring the
interactome before training. Both need the controls used here, in particular
a degree-matched null and a floor estimated inside every subset for which an
increment is claimed.

\appendices

\section{Numerical Verification}
\label{app:numerics}

All structural quantities come from a dense eigendecomposition of the
Laplacian. We check it against two routes that share no code path, and
against the analytic answer where one exists.

The first alternative is the singular value decomposition pseudoinverse.
Since both rely on the same symmetric factorisation, the second uses no
spectral decomposition: with $J$ the all-ones matrix, $L+J/n$ is positive
definite and
\begin{equation}
L^{+} = (L + J/n)^{-1} - J/n ,
\label{eq:bordered}
\end{equation}
solved directly by Cholesky. Two further properties need no reference
implementation: the Moore-Penrose condition $LL^{+}L=L$, and
$L^{+}\mathbf{1}=\mathbf{0}$, which makes every row sum exactly zero.

\begin{table}[h]
\centering
\caption{Largest deviation observed between three routes to $L^{+}$, on the
worst-conditioned graphs of either corpus. Route differences are relative to
the scale of the diagonal.}
\label{tab:numerics}
\footnotesize
\begin{tabular}{lrr}
\toprule
& Peptides & Interactomes \\
& 12 worst cond. & 4 largest \\
\midrule
Nodes & 401--444 & 3195--3480 \\
$\lambda_2$ & $1.5\times 10^{-4}$ & $0.32$--$0.38$ \\
Condition number & $3.5\times 10^{4}$ & $2.1\times 10^{3}$ \\
\midrule
Eigendecomp. vs SVD & $2.5\times 10^{-15}$ & $1.2\times 10^{-14}$ \\
Eigendecomp. vs \eqref{eq:bordered} & $1.4\times 10^{-11}$ & $5.4\times 10^{-14}$ \\
Moore-Penrose residual & $6.2\times 10^{-13}$ & $3.0\times 10^{-14}$ \\
Row sums of $L^{+}$ & $3.1\times 10^{-8}$ & $2.2\times 10^{-12}$ \\
\bottomrule
\end{tabular}
\end{table}

No two routes disagree by more than $1.4\times 10^{-11}$ relative. Row sums
separate the corpora by five orders of magnitude, tracking the spectral gap:
peptides are elongated and their $\lambda_2$ is three orders smaller than an
interactome's.

Exact trees give an absolute bound, since resistance there equals graph
distance in closed form. Over the twelve worst-conditioned trees, $\lambda_2$
from $2.0\times 10^{-4}$ to $3.5\times 10^{-4}$ and 265 to 312 nodes, the
largest relative departure is $1.9\times 10^{-12}$, eight orders of magnitude
below the smallest effect reported in the paper.

\section{Structural Quantities}
\label{app:structure}

The score of \eqref{eq:score} removes degree, so a natural objection is that
the residual is some higher-order degree quantity and that we simply stopped
adding covariates too early. Table~\ref{tab:ladder} answers this by extending
the regression until the covariate itself saturates.

\begin{table}[h]
\centering
\caption{Variance of $\log L^{+}_{ii}$ left unexplained as covariates are
added, averaged over graphs. Coverage is the mean fraction of the network
inside the ball.}
\label{tab:ladder}
\footnotesize
\begin{tabular}{lrrrr}
\toprule
& \multicolumn{2}{c}{Real} & \multicolumn{2}{c}{Null} \\
\cmidrule(lr){2-3}\cmidrule(lr){4-5}
Covariate added & Unexpl. & Incr. & Unexpl. & Incr. \\
\midrule
Degree & $0.935\%$ & --- & $0.675\%$ & --- \\
$+$ neighbour degree & $0.731\%$ & $0.00204$ & $0.596\%$ & $0.00078$ \\
$+$ clustering & $0.710\%$ & $0.00022$ & $0.587\%$ & $0.00009$ \\
$+$ 2-hop ball & $0.303\%$ & $0.00407$ & $0.107\%$ & $0.00481$ \\
$+$ 3-hop ball & $0.260\%$ & $0.00042$ & $0.098\%$ & $0.00009$ \\
\midrule
2-hop coverage & \multicolumn{2}{c}{$0.323$} & \multicolumn{2}{c}{$0.357$} \\
3-hop coverage & \multicolumn{2}{c}{$0.890$} & \multicolumn{2}{c}{$0.937$} \\
\bottomrule
\end{tabular}
\end{table}

The three-hop ball already covers $89\%$ of a real interactome, so the
covariate is close to constant and the sequence has run out rather than been
cut short. Its increment is an order of magnitude below that of the two-hop
ball. The null world settles harder still, at $0.00009$ against $0.00042$,
which is the relevant comparison: what remains unexplained in a real
interactome is not an artefact of an unfinished ladder, since the same ladder
run on a degree-matched graph leaves a third as much. Per graph the figures
after the two-hop ball are $0.303\%$ on average with a range of $0.218\%$ to
$0.863\%$ for real interactomes, against $0.107\%$ and $0.080\%$ to $0.253\%$
for their nulls.

\section{Baselines and Design Choices}
\label{app:baselines}

\subsection{Choice of predictive uncertainty}

$Z_0$ contains predictive entropy rather than Monte Carlo dropout variance.
Table~\ref{tab:baselines} gives the reason. Entropy tracks per-node loss far
more closely than the dropout variance does, and the gap widens with depth.
Margin, the mean distance of the predicted probabilities from one half,
tracks it about as well with the expected negative sign.

\begin{table}[h]
\centering
\caption{Spearman correlation of each predictive uncertainty score with
per-node loss, on validation graphs. A usable score correlates positively;
margin is a confidence score and so correlates negatively.}
\label{tab:baselines}
\footnotesize
\begin{tabular}{lrrr}
\toprule
& \multicolumn{3}{c}{Message-passing depth} \\
\cmidrule(lr){2-4}
& 2 & 4 & 6 \\
\midrule
Entropy vs loss & $+0.844$ & $+0.868$ & $+0.891$ \\
Margin vs loss & $-0.853$ & $-0.884$ & $-0.911$ \\
Dropout variance vs loss & $+0.300$ & $+0.449$ & $+0.529$ \\
\midrule
Dropout variance vs entropy & $+0.492$ & $+0.616$ & $+0.668$ \\
\bottomrule
\end{tabular}
\end{table}

Dropout variance is the weaker score because it peaks where the predicted
probability is near one half, while the highest per-node losses come from
nodes that are confident and wrong. Choosing entropy therefore sets the
hardest available bar for the structural score, not a convenient one.

\subsection{Number of seeds and replicas}

Three independent sources of variation were measured, each as the ratio of
the standard deviation between interactomes to the standard deviation within
one interactome.

\begin{table}[h]
\centering
\caption{Variation between interactomes relative to variation within one.
A large ratio means the sample of graphs, not the number of repetitions,
is the binding constraint.}
\label{tab:noise}
\footnotesize
\begin{tabular}{lrrr}
\toprule
Source & Within & Between & Ratio \\
\midrule
Training seeds & $0.0104$ & $0.0305$ & $2.9$ \\
Rewiring replicas, structural & $0.0125$ & $0.0367$ & $2.9$ \\
Rewiring replicas, downstream & $0.00030$ & $0.00033$ & $1.1$ \\
\bottomrule
\end{tabular}
\end{table}

The first two agree to one decimal place although they arise from unrelated
mechanisms, one from stochastic optimisation and one from random edge swaps.
In both cases genuine heterogeneity between tissue interactomes dominates,
so adding seeds or replicas would not narrow the intervals materially; the
24 graphs are the binding constraint. The third ratio is near one, which
means the rewiring randomness is largely absorbed by the time it reaches the
trained model, and a single replica would have sufficed downstream. We report
this as a measurement rather than assuming it in advance.

\section{Per-Graph Results}
\label{app:pergraph}

\begin{table*}[t]
\centering
\caption{All 24 tissue interactomes, ordered by size. Residual columns give
the variance of $\log L^{+}_{ii}$ left unexplained after the full covariate
ladder, for the real network and for the mean of its three degree-preserving
replicas. $\Delta R^{2}$ is the incremental variance in per-node loss under
$Z_0^{+}$ at six layers, averaged over five seeds. The permutation floor,
$+0.00021$, is given for comparison.}
\label{tab:pergraph}
\footnotesize
\begin{tabular}{lrrrrrrr}
\toprule
Split and index & Nodes & Mean degree & $D_{\deg}$ & Resid. real (\%) &
Resid. null (\%) & Ratio & $\Delta R^{2}$ \\
\midrule
train[8] & 591 & 13.04 & 0.0851 & 0.863 & 0.253 & 3.41 & $0.00105$ \\
train[5] & 1021 & 17.84 & 0.0571 & 0.427 & 0.143 & 2.99 & $0.00280$ \\
train[1] & 1377 & 21.57 & 0.0513 & 0.337 & 0.141 & 2.38 & $0.00024$ \\
train[0] & 1546 & 20.90 & 0.0491 & 0.554 & 0.143 & 3.89 & $0.00014$ \\
train[4] & 1578 & 22.92 & 0.0449 & 0.284 & 0.107 & 2.66 & $0.00011$ \\
train[12] & 1819 & 25.16 & 0.0418 & 0.259 & 0.101 & 2.55 & $0.00473$ \\
train[6] & 1823 & 24.32 & 0.0425 & 0.295 & 0.118 & 2.49 & $0.00213$ \\
train[11] & 1878 & 24.64 & 0.0429 & 0.228 & 0.118 & 1.93 & $0.00011$ \\
train[2] & 2263 & 26.36 & 0.0414 & 0.269 & 0.102 & 2.65 & $0.00033$ \\
test[1] & 2300 & 26.66 & 0.0390 & 0.255 & 0.096 & 2.66 & $0.00025$ \\
train[15] & 2326 & 25.74 & 0.0411 & 0.254 & 0.099 & 2.58 & $0.00040$ \\
train[3] & 2339 & 27.97 & 0.0379 & 0.259 & 0.106 & 2.45 & $0.00071$ \\
train[10] & 2401 & 26.75 & 0.0400 & 0.233 & 0.088 & 2.65 & $0.00087$ \\
train[7] & 2488 & 28.29 & 0.0373 & 0.303 & 0.088 & 3.46 & $0.00089$ \\
train[16] & 2650 & 29.08 & 0.0358 & 0.291 & 0.089 & 3.28 & $0.00097$ \\
train[14] & 2794 & 30.54 & 0.0344 & 0.226 & 0.080 & 2.82 & $0.00135$ \\
train[17] & 2815 & 30.38 & 0.0347 & 0.231 & 0.087 & 2.65 & $0.00321$ \\
train[19] & 3021 & 30.23 & 0.0353 & 0.250 & 0.085 & 2.94 & $0.00257$ \\
train[18] & 3163 & 29.77 & 0.0359 & 0.227 & 0.099 & 2.30 & $0.00086$ \\
test[0] & 3194 & 31.51 & 0.0332 & 0.222 & 0.081 & 2.75 & $0.00063$ \\
val[0] & 3195 & 30.50 & 0.0344 & 0.242 & 0.086 & 2.82 & $0.00092$ \\
val[1] & 3284 & 30.90 & 0.0341 & 0.218 & 0.085 & 2.58 & $0.00140$ \\
train[9] & 3312 & 32.06 & 0.0337 & 0.243 & 0.085 & 2.85 & $0.00092$ \\
train[13] & 3480 & 30.68 & 0.0347 & 0.300 & 0.085 & 3.55 & $0.00005$ \\
\midrule
Mean & 2361 & 26.57 & 0.0416 & 0.303 & 0.107 & 2.80 & $0.00115$ \\
\bottomrule
\end{tabular}
\end{table*}

Table~\ref{tab:pergraph} gives every number the aggregate claims rest on.
Three patterns are worth separating.

The structural result is uniform. The real interactome leaves more
unexplained variance than its degree-matched null in all 24 cases, and the
ratio never falls below $1.93$ nor rises above $3.89$. There is no graph for
which the ordering is close.

The size trend in $D_{\deg}$ is monotone across nearly the whole range,
from $0.085$ in the smallest network to $0.033$ in the largest, which is what
the Spearman correlation of $-0.955$ reports.

The risk result is uniform in sign but not in magnitude. $\Delta R^{2}$ is
positive in every interactome, but it spans $0.00005$ to $0.00473$, a factor
of nearly one hundred, and five networks fall below the permutation floor of
$+0.00021$. The aggregate interval excludes zero because the effect is
consistent in direction, not because it is strong in any single tissue. Nor
does the spread follow network size: the smallest and the largest interactome
sit at opposite ends of it. Baseline difficulty varies as widely, with the
$R^{2}$ of $Z_0$ ranging from $0.41$ to $0.80$ across tissues, so the
consistency in sign holds across quite different regimes of predictability.

\section{A Mechanism Test That Fails}
\label{app:receptive}

If over-squashing acted node by node, proteins whose two-hop neighbourhood
covers less of the network should show a stronger effect, since they depend
on more distant information. Splitting nodes by that coverage appears to
confirm it. Across coverage quartiles the corrected increment falls
monotonically, from $+0.00191$ in the least covered quartile to $+0.00038$ in
the most covered. Mean degree over the same quartiles runs from $4.5$ to
$69.1$, and the Spearman correlation between degree and coverage is $+0.917$,
so the quartiles are largely re-slicing degree, which $Z_0^{+}$ already
contains.

Blocking on degree first removes the effect. Within each degree block the
nodes are split again by coverage, and the low-coverage half is compared
against the high-coverage half of the same block, as
Table~\ref{tab:receptive} shows.

\begin{table}[h]
\centering
\caption{Corrected increment within degree blocks, split by two-hop coverage.
Each increment has its own permutation floor subtracted, since that floor
grows as the subset shrinks.}
\label{tab:receptive}
\footnotesize
\begin{tabular}{lrrrr}
\toprule
Degree block & Coverage & Mean deg. & Nodes & Corrected $\Delta R^{2}$ \\
\midrule
Low & $0.052$ & $3.3$ & 414 & $+0.00232$ \\
Low & $0.202$ & $5.8$ & 413 & $+0.00231$ \\
Middle & $0.231$ & $13.6$ & 381 & $+0.00022$ \\
Middle & $0.387$ & $17.7$ & 378 & $-0.00017$ \\
High & $0.444$ & $35.4$ & 388 & $+0.00011$ \\
High & $0.653$ & $86.5$ & 386 & $+0.00094$ \\
\bottomrule
\end{tabular}
\end{table}

Paired within blocks, low coverage minus high coverage gives $+0.000014$
$[-0.00228, +0.00297]$ in the low-degree block, $+0.00039$
$[-0.00044, +0.00134]$ in the middle, and $-0.00083$ $[-0.00216, +0.00056]$
in the high, positive in 10, 11 and 10 of 24 graphs respectively. Pooled
across blocks the difference is $-0.00014$ $[-0.00110, +0.00102]$, positive
in 31 of 72 cells. A continuous alternative that avoids binning, an
interaction between the score and log coverage, gives $+0.00057$
$[-0.00026, +0.00126]$ and is negative in only 5 of 24 graphs, the opposite
sign to the prediction.

Both routes agree that coverage carries nothing once degree is fixed. The
mechanism evidence in this work is therefore model-level, resting on the
depth trend, and not node-level. This is consistent with the account of
\cite{digiovanni2023}, under which depth does not relieve over-squashing but
allows vanishing gradients to dominate it, since that is a property of the
trained model rather than of any individual node.

\section{Derivations}
\label{app:derivations}

\subsection{The inverse-degree decomposition}

Write $\epsilon_{ij} = R_{\mathrm{eff}}(i,j) - 1/d_i - 1/d_j$ for the
departure of a pair from the inverse-degree limit, and
$S_G = \sum_{j \in V} 1/d_j$. Since $R_{\mathrm{eff}}(i,i)=0$,
\begin{equation}
\sum_{j \neq i} \Big(\tfrac{1}{d_i} + \tfrac{1}{d_j} + \epsilon_{ij}\Big)
= \frac{n-1}{d_i} + \Big(S_G - \frac{1}{d_i}\Big) + \sum_{j \neq i}
\epsilon_{ij} .
\end{equation}
Equating this with \eqref{eq:nodesum} and dividing by $n$ gives
\begin{equation}
L^{+}_{ii} = \frac{n-2}{n\,d_i} + C_G
+ \frac{1}{n}\sum_{j \neq i} \epsilon_{ij} ,
\qquad C_G = \frac{S_G - \mathrm{tr}(L^{+})}{n} ,
\end{equation}
where $C_G$ depends on the graph but not on the node. The theory-aligned
score \eqref{eq:cer} is therefore the node-averaged departure from the limit,
up to that constant. This is an algebraic rearrangement rather than a result;
we state it because it makes explicit what a fixed-coefficient correction
removes and what it leaves behind.

\subsection{The shorting bound}

For non-adjacent $i$ and $j$, short every other node of $G$ into a single
supernode $S$. Every edge at $i$ then terminates at $S$, since none of its
neighbours is $j$, so the resistance from $i$ to $S$ is $1/d_i$; likewise
$1/d_j$ from $j$ to $S$. The two are in series, giving $1/d_i + 1/d_j$ in the
shorted network. Rayleigh monotonicity states that shorting cannot increase
an effective resistance, so
\begin{equation}
R_{\mathrm{eff}}(i,j) \geq \frac{1}{d_i} + \frac{1}{d_j} ,
\qquad i \nsim j ,
\end{equation}
and $\epsilon_{ij} \geq 0$ on such pairs. The argument is classical. The
consequence for us is that the numerator of \eqref{eq:ddeg} is a non-negative
excess rather than a signed residual, which is why we restrict that quantity
to non-adjacent pairs. Empirically the bound is not merely satisfied but
attained: the smallest $\epsilon_{ij}$ over non-adjacent pairs is
$-1.1 \times 10^{-13}$ across all 24 interactomes and $-7.5 \times 10^{-14}$
across their nulls, both zero to machine precision, with equality reached by
pairs of low-degree nodes.

\end{document}